\documentclass[journal, 12pt]{IEEEtran}
\onecolumn
\usepackage{setspace}
\usepackage{amsmath}
\usepackage{amsthm}   
\usepackage{amssymb} 
\usepackage[bottom,hang,flushmargin]{footmisc}
\usepackage{graphicx}
\usepackage{caption}
\usepackage{multirow} 
\renewcommand{\footnoterule}{%
  \kern -3pt
  \hrule width 0.4\textwidth height 0.4pt
  \kern 2.6pt
}

\begin{document}

\title{Unveiling the Power of Wavelets: A Wavelet-based Kolmogorov-Arnold Network for Hyperspectral Image Classification}

\author{Seyd Teymoor Seydi,~\IEEEmembership{Member,~IEEE} \thanks{ seydi.teymoor@ut.ac.ir} }

\author{Seyd Teymoor Seydi,~\IEEEmembership{Member,~IEEE} \thanks{ S.T. Seydi (seydi.teymoor@ut.ac.ir) School of Surveying and Geospatial Engineering, College of Engineering, University of Tehran, Tehran, Iran}, Zavareh Bozorgasl,~\IEEEmembership{Member,~IEEE}, Hao~Chen,~\IEEEmembership{Member,~IEEE}  \thanks{Z. Bozorgasl (zavarehbozorgasl@boisestate.edu) and H. Chen (haochen@boisestate.edu) are with the Department of Electrical and Computer Engineering, Boise State University, Boise,
ID, 83712.} }
\maketitle

\begin{abstract}

\noindent Hyperspectral image classification is a crucial but challenging task due to the high dimensionality and complex spatial-spectral correlations inherent in hyperspectral data. This paper employs Wavelet-based Kolmogorov-Arnold Network (wav-kan) \cite{wav-kan} architecture tailored for efficient modeling of these intricate dependencies. Inspired by the Kolmogorov-Arnold representation theorem, Wav-KAN incorporates wavelet functions as learnable activation functions, enabling non-linear mapping of the input spectral signatures. The wavelet-based activation allows Wav-KAN to effectively capture multi-scale spatial and spectral patterns through dilations and translations. Experimental evaluation on three benchmark hyperspectral datasets (Salinas, Pavia, Indian Pines) demonstrates the superior performance of Wav-KAN compared to traditional multilayer perceptrons (MLPs) and the recently proposed Spline-based KAN (Spline-KAN) model \cite{kan}. Across the datasets, Wav-KAN achieved an average overall accuracy of 92.62\% and kappa coefficient of 0.9157, outperforming Spline-KAN (avg. accuracy 89.85\%, kappa 0.8793) and MLP (avg. accuracy 77.69\%, kappa 0.7119). On the challenging Indian Pines dataset, Wav-KAN's significant performance gains (85.54\% accuracy, 0.8348 kappa) over Spline-KAN (77.31\%, 0.7395) and MLP (35.13\%, 0.2984) underscore its ability to model complex spectral-spatial dependencies effectively. The competitive results of Spline-KAN, especially on Salinas, highlight the potential of theory-driven kernels like splines and wavelets in hyperspectral classification tasks. Wav-KAN's wavelet-based architecture enables interpretable, non-linear mapping while capturing multi-scale patterns, demonstrating the advantages of incorporating representation theorems into deep learning models for enhanced performance on high-dimensional, correlated data like hyperspectral imagery.
\textbf{In this work,} we are: (1) conducting more experiments on additional hyperspectral datasets (Pavia University, WHU-Hi, and Urban Hyperspectral Image) to further validate the generalizability of Wav-KAN; (2) developing a multiresolution Wav-KAN architecture to capture scale-invariant features; (3) analyzing the effect of dimensional reduction techniques on classification performance; (4) exploring optimization methods for tuning the hyperparameters of KAN models; and (5) comparing Wav-KAN with other state-of-the-art models in hyperspectral image classification.
\end{abstract}


\IEEEpeerreviewmaketitle

\section{Introduction}

Hyperspectral imaging is a remote sensing technique that captures the reflectance or emission of light from an object across a wide range of contiguous spectral bands, typically ranging from visible to infrared regions of the electromagnetic spectrum \cite{eismann2012hyperspectral}. Unlike traditional RGB or multispectral images, which have a limited number of broad bands, hyperspectral images consist of hundreds of narrow, closely spaced spectral bands, providing detailed spectral information for each pixel in the image. This rich spectral information enables the identification and discrimination of various materials, surface features, and objects based on their unique spectral signatures \cite{bioucas2013hyperspectral}.

The high spectral resolution of hyperspectral data allows researchers to precisely characterize the electromagnetic spectrum of an object. Mathematically, a hyperspectral image can be represented as a 3D cube, with two spatial dimensions and one spectral dimension. However, most hyperspectral sensors have a low spatial resolution, so each pixel corresponds to a mixture of materials observed over a surface area. This spectral mixing poses challenges for accurate classification and requires advanced techniques to handle the mixed pixel problem, such as spectral unmixing \cite{bioucas2012hyperspectral}.

Deep learning techniques have shown remarkable success in hyperspectral image classification. These models can be categorized into three main groups based on their learning approach: Supervised Learning, Semi-Supervised Learning, and Unsupervised Learning. 
In supervised learning, deep neural networks are trained using labeled data, where both input (hyperspectral image pixels) and corresponding output labels (classes) are provided during the training process. Examples of supervised models include Convolutional Neural Networks (CNNs), Recurrent Neural Networks (RNNs), and fully connected deep neural networks. These models learn to extract discriminative features from the input data and map them to the desired output labels through a hierarchical network architecture \cite{chen2016deep}. CNNs are particularly well-suited for hyperspectral data due to their ability to capture local spatial and spectral patterns through convolutional and pooling operations. The convolutional layers apply learnable filters to the input data, generating feature maps that capture different patterns. These feature maps are then subsampled by pooling layers, reducing their spatial dimensions while retaining the most important features. This process is repeated in multiple layers, with each layer learning increasingly abstract and complex representations \cite{hu2015deep}. RNNs, on the other hand, are designed to handle sequential data and can model long-range dependencies in hyperspectral pixels by treating them as a sequence of spectral values. RNNs have been used for hyperspectral classification by incorporating recurrent layers that can capture spectral-spatial contexts \cite{mou2017deep}.

Hyperspectral datasets pose several challenges for deep learning models due to their inherent complexity. Hyperspectral images have hundreds of spectral bands, resulting in a high-dimensional feature space. This high dimensionality can lead to the "curse of dimensionality," where the number of training samples required for accurate modeling increases exponentially with the number of dimensions \cite{hughes1968mean}. Deep learning models may struggle to effectively learn and generalize from high-dimensional data, especially when the available training samples are limited.
Hyperspectral images exhibit complex spatial-spectral correlations, where neighboring pixels and spectral bands are often highly correlated. Capturing and modeling these correlations is crucial for accurate classification but can be challenging for deep learning models \cite{fang2015spectral}. Traditional models may treat spatial and spectral information separately, failing to capture the intricate relationships between them.
Due to the relatively coarse spatial resolution of hyperspectral sensors, a single pixel may contain a mixture of materials or spectra, known as mixed pixels \cite{somers2011endmember}. This spectral mixing poses challenges for accurate classification, as the observed spectrum is a combination of multiple endmembers (pure material spectra). Additionally, spectral variability within the same material class can arise due to factors like illumination conditions, atmospheric effects, and sensor noise, further complicating the classification task \cite{chen2011spectral}.
Despite these challenges, deep learning techniques have shown promising results in hyperspectral image classification, often outperforming traditional machine learning methods. Ongoing research focuses on developing more robust and efficient deep learning models tailored specifically for hyperspectral data, along with strategies for dealing with limited labeled data, high dimensionality, and spatial-spectral correlations.

Kolmogorov-Arnold Networks (KANs), grounded in the Kolmogorov-Arnold representation theorem \cite{kan_old}, provide notable benefits including enhanced interpretability and accuracy \cite{kan}. These networks feature univariate learnable activation functions on their edges, with nodes summing these activation functions. Most previous studies on KANs have concentrated on depth-2 representations, with the exception of Liu et al. \cite{kan}, who expanded KANs to support arbitrary widths and depths, a variant known as B-Spline KAN.

Within the realm of Universal Approximation Theory (UAT), wavelets have been investigated in neural networks. This study applies wavelets to the Kolmogorov-Arnold representation theorem for neural networks with arbitrary widths and depths \cite{wav-kan}, targeting hyperspectral image classification through both continuous wavelet transform (CWT) and discrete wavelet transform (DWT).

Section \ref{sec:methods} discusses the methodologies that we applied to our datasets. Section \ref{sec:dataset1} introduces benchmark hyperspectral datasets inclduing Indian Pines, Pavia, and Salinas \footnote{We will add more simulations in next versions.}.  Section \ref{sec:exp} presents Experiment and Results. The comparison of Wav-KAN, Spline-KAN and MLPs is given. Finally, Section \ref{sec:conclusion} concludes the paper.

\section{Methods}
\label{sec:methods}
Wavelet Kolmogorov-Arnold Networks (Wav-KAN) offer several advantages over Spline Kolmogorov-Arnold Networks (Spline-KAN) and traditional multilayer perceptrons (MLPs). First and foremost, Wav-KAN enhances interpretability and performance by leveraging wavelet functions within the Kolmogorov-Arnold network structure. This allows Wav-KAN to efficiently capture both high-frequency and low-frequency components of the input data. Wavelets, with their ability to perform multiresolution analysis, ensure that the network can isolate significant patterns while discarding irrelevant noise, providing a more accurate and nuanced representation of the original data without the pitfalls of noise overfitting. This feature makes Wav-KAN superior in terms of interpretability and robustness compared to Spline-KAN and MLPs.\\

In terms of computational efficiency and training speed, Wav-KAN also holds an edge. Unlike Spline-KAN, which requires smooth functions and grid spaces for better performance—often resulting in cumbersome and computationally expensive operations—Wav-KAN utilizes the inherent scaling properties of wavelets. This reduces the need for additional terms in activation functions and speeds up training. Additionally, the wavelet-based approach in Wav-KAN obviates the need for recalculating previous steps when capturing more details, as discrete wavelet transform (DWT) can efficiently combine local detailed information with broader trends. This results in faster and more efficient training compared to both Spline-KAN and MLPs.\\

Moreover, Wav-KAN benefits from having fewer parameters while maintaining high performance. By employing wavelets, which can capture both low-frequency and high-frequency functions, Wav-KAN networks require fewer parameters than their Spline-KAN and MLP counterparts. This not only enhances the model's generalization ability but also leads to significant improvements in accuracy and robustness. The adaptive nature of wavelets allows Wav-KAN to conform to the data structure, akin to how water conforms to the shape of its container, thus achieving higher accuracy and robustness compared to Spline-KAN and MLPs. Overall, Wav-KAN provides a powerful and efficient neural network model with applications across various fields.
\section{Dataset}
\label{sec:dataset1}
In this study we used three benchmark hyperspectral datasets. 
\begin{figure}[ht]
   \centering
   \begin{tabular}{cc}
       \includegraphics[width=0.45\linewidth, height=7cm]{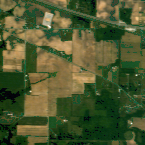} & \includegraphics[width=0.45\linewidth, height=7cm]{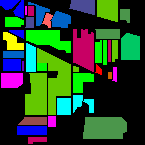} \\
       (a) Color composite Indian Pine dataset & (b) Ground Truth \\
       \includegraphics[width=0.45\linewidth, height=9cm]{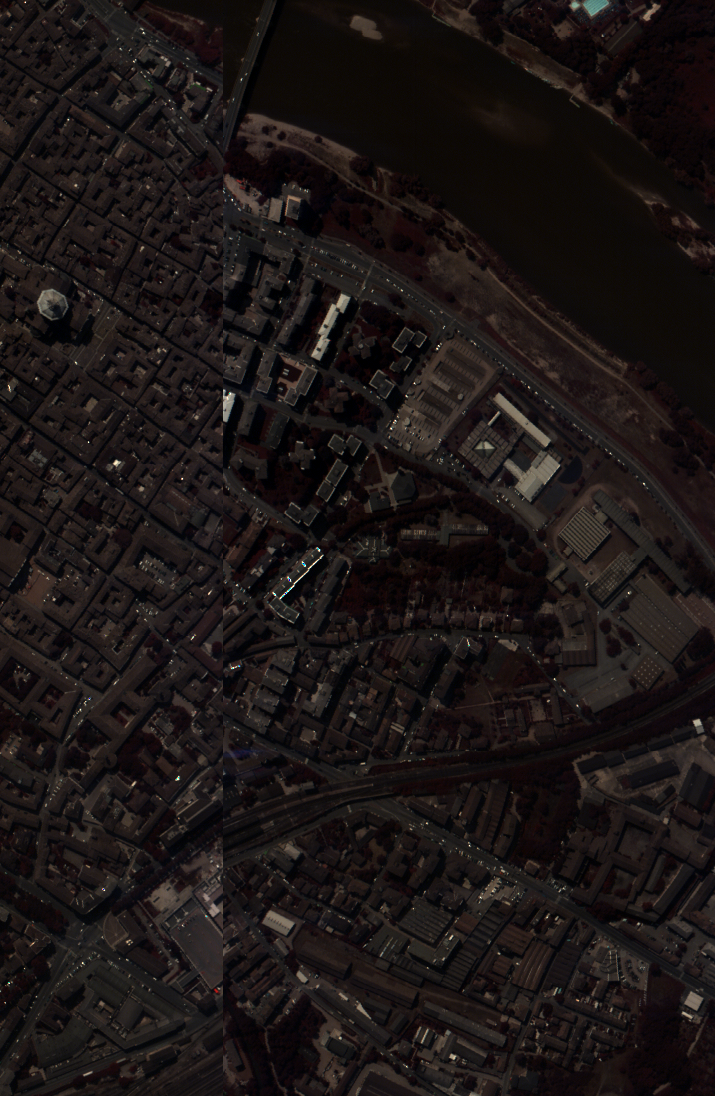} & \includegraphics[width=0.45\linewidth, height=9cm]{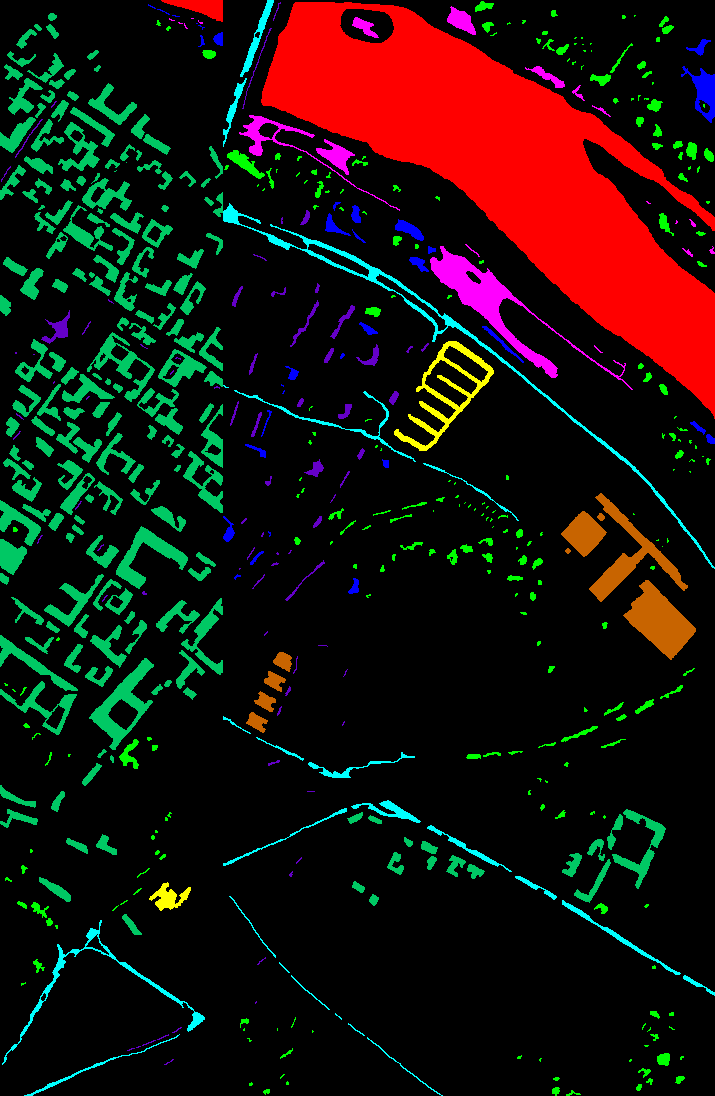} \\
       (c) Color composite Pavia dataset & (d) Ground Truth\\
       \includegraphics[width=0.45\linewidth, height=7cm]{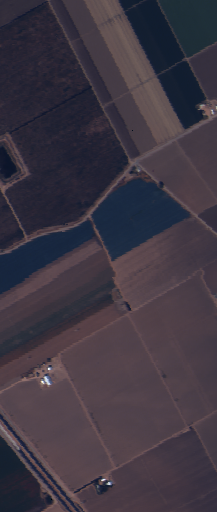} & \includegraphics[width=0.45\linewidth, height=7cm]{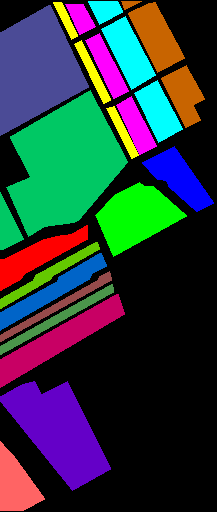} \\
       (e) Color composite Salinas dataset & (f) Ground Truth \\
   \end{tabular}
   \captionsetup{justification=centering}
   \caption{Incorporate hyperspectral dataset in this study}
   \label{fig:StudyArea}
\end{figure}

\subsection{Indian Pines }

The Indian Pines dataset, acquired by the Airborne Visible/Infrared Imaging Spectrometer (AVIRIS) sensor on June 12, 1992, covers an agricultural area in Northwestern Indiana, USA. The dataset consists of 224 spectral bands ranging from 400-2500 nm, with a spatial resolution of 20 m per pixel. The scene, comprising 145 x 145 pixels Figure \ref{fig:StudyArea}a, contains 16 distinct land cover classes Figure \ref{fig:StudyArea}b, primarily focusing on different crop types and vegetation species \cite{PURR1947}.

\subsection{Pavia }
The Pavia Centre dataset is a hyperspectral image acquired by the ROSIS sensor during a flight campaign over Pavia, northern Italy. It consists of 102 spectral bands with a spatial dimension of 1096 x 715 pixels Figure \ref{fig:StudyArea}c. The geometric resolution of the dataset is 1.3 meters. The ground truth data for the Pavia Centre scene differentiates 9 classes Figure \ref{fig:StudyArea}d \cite{paviadataset}. 

\subsection{Salinas }
The Salinas dataset is a hyperspectral image acquired over the Salinas Valley, California. The study area covers 512 x 217 samples with a spatial resolution of 3.7-meter pixels Figure \ref{fig:StudyArea}e. The dataset originally consisted of 224 spectral bands acquired by the AVIRIS sensor, but 20 water absorption bands were removed, leaving 204 bands. The ground truth data contains 16 classes representing different land cover types in the region Figure \ref{fig:StudyArea}f \cite{melgani2004classification}.

\section{Experiment and Results}
\label{sec:exp}
\subsection{Visual Analysis}

\begin{figure}[ht]
   \centering
   \begin{tabular}{cc}
       \includegraphics[width=0.45\linewidth, height=9cm]{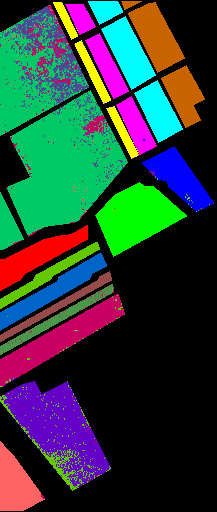} & \includegraphics[width=0.45\linewidth, height=9cm]{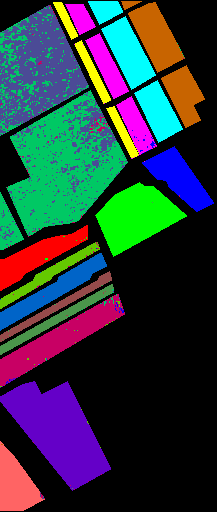} \\
       (a) & (b) \\
       \includegraphics[width=0.45\linewidth, height=9cm]{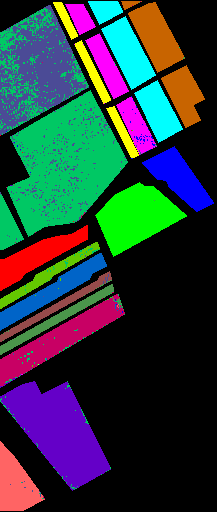} & \includegraphics[width=0.45\linewidth, height=9cm]{figures/Salinas_gt.png} \\
       (c) & (d) \\
   \end{tabular}
   \captionsetup{justification=centering}
   \caption{Result of hyperspectral classification for Salinas dataset. (a) MLP, (b) Spline-KAN, (c) Wav-KAN, (d)  Ground Truth}
   \label{fig:Salinas}
\end{figure}

The provided  Figure \ref{fig:Salinas} showcases the results of hyperspectral classification for the Salinas. A qualitative evaluation indicates that all three methodologies exhibit robust performance in discriminating the major land cover classes, as demonstrated by the strong spatial and spectral correspondence between the classified images and the ground truth. Nevertheless, close inspection reveals minor discrepancies, suggesting potential misclassifications or class confusion. The MLP classifier appears to have a slightly higher tendency for misclassification (Figure \ref{fig:Salinas}a), especially in border regions between classes, when compared to the Spline-KAN (Figure \ref{fig:Salinas}b) and Wav-KAN approaches (Figure \ref{fig:Salinas}c).

\begin{figure}[ht]
   \centering
   \begin{tabular}{cc}
       \includegraphics[width=0.45\linewidth, height=9cm]{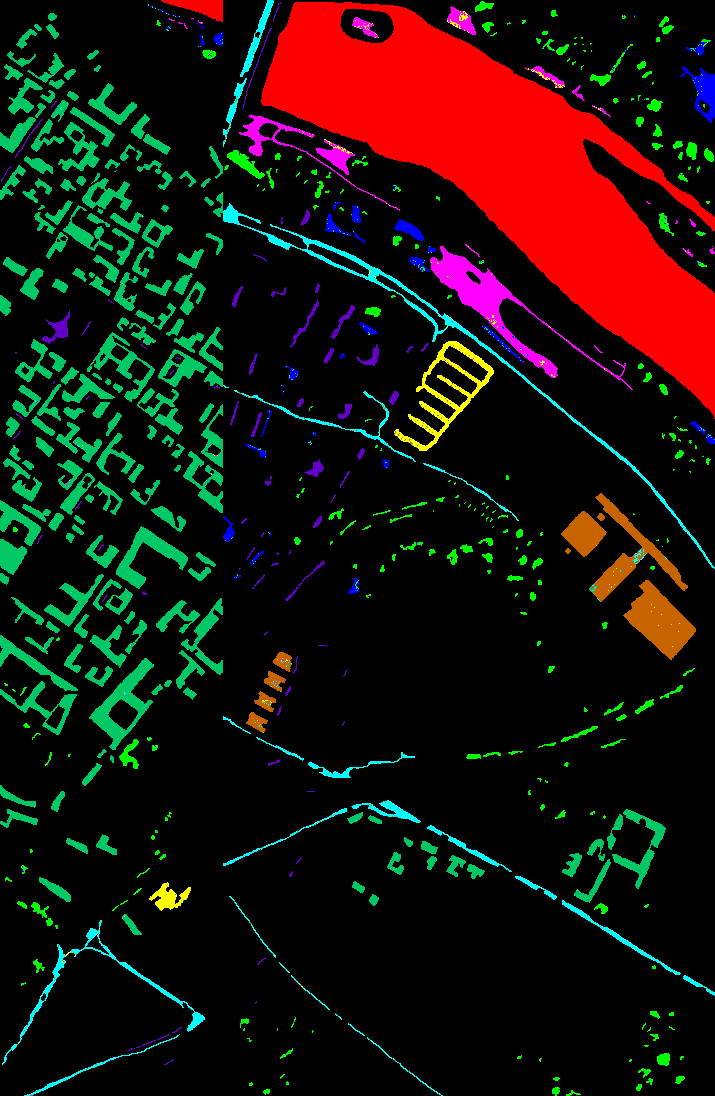} & \includegraphics[width=0.45\linewidth, height=9cm]{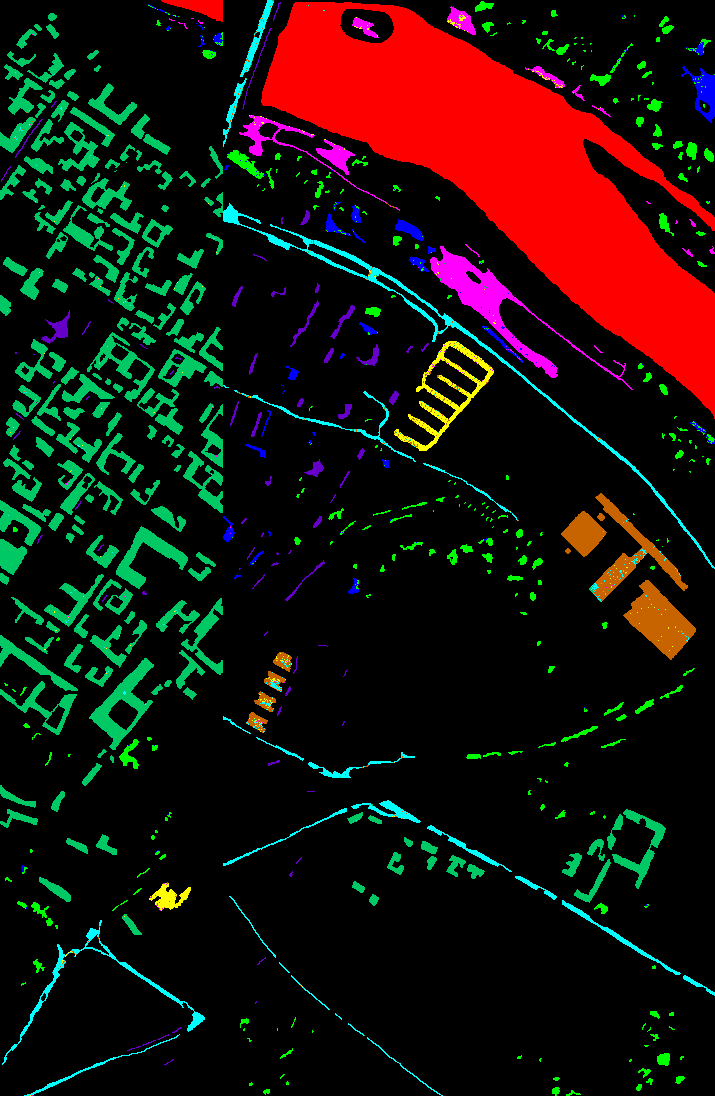} \\
       (a) & (b) \\
       \includegraphics[width=0.45\linewidth, height=9cm]{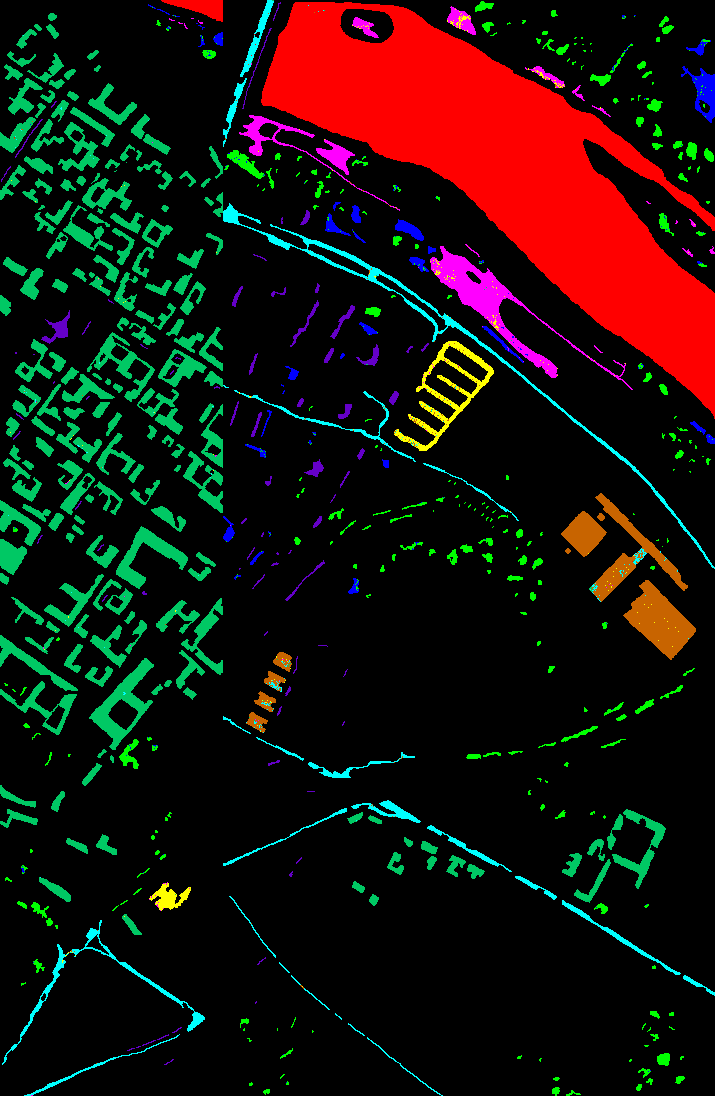} & \includegraphics[width=0.45\linewidth, height=9cm]{figures/Pavia_gt.png} \\
       (c) & (d) \\
   \end{tabular}
   \captionsetup{justification=centering}
   \caption{Result of hyperspectral classification for Pavia dataset. (a)  MLP, (b) Spline-KAN, (c) Wav-KAN, (d) Ground Truth}
   \label{fig:Pavia}
\end{figure}

Based on the visual assessment of the classification results presented in the Figure \ref{fig:Pavia}, the MLP Figure \ref{fig:Pavia}a approach appears to achieve the best overall performance among the three methods compared for the Pavia dataset. The MLP classifier demonstrates a higher degree of spatial and spectral consistency with the ground truth data, suggesting more accurate discrimination of urban land cover classes. The Wav-KAN (Figure \ref{fig:Pavia}c) exhibits better performance than the Spline-KAN (Figure \ref{fig:Pavia}b), as evidenced by fewer misclassifications and a more coherent representation of complex spatial patterns. 

\begin{figure}[ht]
    \centering
    \begin{tabular}{cc}
        \includegraphics[width=0.45\linewidth]{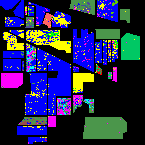} & \includegraphics[width=0.45\linewidth]{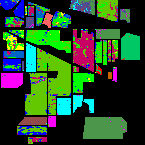} \\
        (a) & (b) \\
        \includegraphics[width=0.45\linewidth]{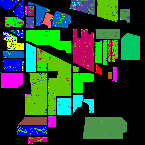} & \includegraphics[width=0.45\linewidth]{figures/Indian_pines_gt.png} \\
        (c) & (d) \\
    \end{tabular}
    \captionsetup{justification=centering}
    \caption{(a) Indian Pines MLP, (b) Indian Pines Spline-KAN, (c) Indian Pines Wav-KAN, (d) Indian Pines Ground Truth}
    \label{fig:Indipain}
\end{figure}

The MLP model Figure \ref{fig:Indipain}(a) displays significant noise, fragmentation, and misclassifications, with small, scattered regions and poorly defined edges, indicating its inability to effectively capture the intricate spectral-spatial patterns. The Spline-KAN model  Figure \ref{fig:Indipain}(b) demonstrates an improvement, with larger, more homogeneous regions and reduced noise, but still suffers from some misclassifications and jagged edges. In contrast, the Wav-KAN model  Figure \ref{fig:Indipain}(c) and the unspecified model  Figure \ref{fig:Indipain}(d) produce visually superior results, characterized by well-defined, smooth edges, minimal noise or fragmentation, and consistent, large homogeneous regions, suggesting effective capture of the complex spectral-spatial information. These models exhibit fewer misclassifications and better delineation of boundaries between different classes, contributing to their enhanced visual quality and classification accuracy on this challenging hyperspectral dataset.

\subsection{Numerical Results}

Table \ref{tab:results} presents the results of hyperspectral classification using three different models: Spline-KAN, Wav-KAN, and MLP, on three datasets: Salinas, Pavia, and Indian Pine. The performance metrics used for evaluation are Overall Accuracy (OA) and Kappa coefficient ($\kappa$).

\textbf{Salinas Dataset:}
On the Salinas dataset, the Wav-KAN model demonstrates superior performance, achieving the highest OA of 0.9341 and $\kappa$ of 0.9264. This indicates that the Wav-KAN model effectively captures the spectral and spatial features of the Salinas dataset, resulting in accurate classification. The Spline-KAN model closely follows with an OA of 0.9261 and $\kappa$ of 0.9178, suggesting that the spline-based kernel also provides a good representation of the data. The MLP model, however, lags behind with an OA of 0.8655 and $\kappa$ of 0.8499, indicating that the non-linear mapping of the MLP may not be as effective in capturing the complex relationships within the Salinas dataset.

\textbf{Pavia Dataset:}
For the Pavia dataset, all three models exhibit excellent performance, with OA and $\kappa$ values surpassing 0.98. The MLP model slightly outperforms the others, achieving an OA of 0.9910 and $\kappa$ of 0.9873. This suggests that the non-linear mapping of the MLP is particularly suitable for the Pavia dataset, effectively capturing the underlying patterns. The Wav-KAN and Spline-KAN models also perform remarkably well, with OA values of 0.9901 and 0.9863, and $\kappa$ values of 0.9860 and 0.9806, respectively. The high performance across all models indicates that the Pavia dataset may have more distinguishable features that are easier to classify.

\textbf{Indian Pine Dataset:}
The Indian Pine dataset presents a more challenging classification task, as evidenced by the lower OA and $\kappa$ values compared to the other datasets. The Wav-KAN model significantly outperforms the other models, achieving an OA of 0.8554 and $\kappa$ of 0.8348. This suggests that the wavelet-based kernel in the Wav-KAN model is more effective in capturing the complex spectral and spatial dependencies in the Indian Pine dataset. The Spline-KAN model follows with an OA of 0.7731 and $\kappa$ of 0.7395, indicating that the spline-based kernel still provides a reasonable representation of the data. However, the MLP model struggles on this dataset, with a substantially lower OA of 0.3513 and $\kappa$ of 0.2984. This poor performance suggests that the non-linear mapping of the MLP may not be suitable for the intricate characteristics of the Indian Pine dataset.

\begin{table}[]
\centering
\caption{Results on Different Datasets}
\label{tab:results}
\begin{tabular}{llll}
\hline
Dataset & Model & Overall Accuracy & Kappa \\ \hline
\multirow{3}{*}{Salinas} & Spline-KAN & 0.9261 & 0.9178 \\
 &  Wav-KAN & \textbf{0.9341} & \textbf{0.9264} \\
 & MLP & 0.8655 & 0.8499 \\ \hline
\multirow{3}{*}{Pavia} & Spline-KAN & 0.9863 & 0.9806 \\
 &   Wav-KAN & \ 0.9901& \ 0.9860 \\
 & \ MLP & \textbf{0.9910} & \textbf{0.9873} \\ \hline
\multirow{3}{*}{Indian Pine} & Spline-KAN & 0.7731 & 0.7395 \\
 &  Wav-KAN & \textbf{0.8554} & \textbf{0.8348} \\
 & MLP & 0.3513 & 0.2984 \\ \hline
\end{tabular}
\end{table}

\section{Conclusion}
\label{sec:conclusion}

This study evaluated the Spline-KAN, Wav-KAN, and MLP models for hyperspectral image classification on the Salinas, Pavia, and Indian Pines datasets. The Wav-KAN model performed best on Salinas and Indian Pines, indicating its effectiveness in capturing complex spectral-spatial features. The MLP excelled on Pavia, suggesting its suitability for urban areas. On Salinas, Spline-KAN closely followed Wav-KAN, showing the spline kernel's potential. However, MLP struggled with Indian Pines, highlighting its limitations for intricate datasets. Overall, Wav-KAN emerged as the most robust and versatile approach, but MLP's strengths on Pavia suggest scenario-specific advantages. These findings underscore the importance of model selection based on dataset characteristics and the need for advanced techniques to handle diverse hyperspectral data effectively. Exploring ensemble or hybrid models combining multiple approaches could further improve classification accuracy and robustness.

\ifCLASSOPTIONcaptionsoff
  \newpage
\fi

\bibliographystyle{IEEEtran}
\bibliography{references}

\end{document}